\documentclass{article}
\usepackage[utf8]{inputenc}
\usepackage[a4paper, margin=1.5in]{geometry}
\usepackage[dvipsnames]{xcolor}
\usepackage{fancyhdr}
\usepackage{authblk}
\usepackage{lipsum}
\usepackage{float}
\usepackage{enumitem}
\usepackage{comment}
\usepackage[font=small,labelfont=bf]{caption}
\usepackage[most]{tcolorbox}
\tcbuselibrary{breakable}
\usepackage[natbib,
backend=biber,
bibstyle=apa,
citestyle=numeric-comp, 
sorting=none, 
]{biblatex}
\AtBeginBibliography{\small}

\usepackage{verbatim}

\newcommand{\detailtexcount}[1]{%
  \immediate\write18{texcount -merge -sum -q #1.tex > #1.wcdetail }%
  \verbatiminput{#1.wcdetail}%
}

\usepackage{subfiles} 

\fancypagestyle{plain}{}

\title{\textbf{Deep Learning and the \\Global Workspace Theory}}
\author[1, 2]{Rufin VanRullen}
\author[3]{Ryota Kanai}
\affil[1]{CerCo, CNRS UMR5549, Toulouse, France}
\affil[2]{ANITI, Universit\'e de Toulouse, France}
\affil[3]{Araya Inc, Tokyo. Japan}
\date{}                     
\defbibenvironment{bibliography}
{\enumerate{}
{\setlength{\leftmargin}{\bibhang}%
\setlength{\itemindent}{-\leftmargin}%
\setlength{\itemsep}{\bibitemsep}%
\setlength{\parsep}{\bibparsep}}}
{\endenumerate}
{\item}

\DeclareUnicodeCharacter{0301}{\'{e}}
\begin{document}
\captionsetup[figure]{labelfont={bf},labelformat={default},labelsep=period,name={Fig.}}
\maketitle

\section*{Abstract}
Recent advances in deep learning have allowed Artificial Intelligence (AI) to reach near human-level performance in many sensory, perceptual, linguistic or cognitive tasks. There is a growing need, however, for novel, brain-inspired cognitive architectures. The Global Workspace theory refers to a large-scale system integrating and distributing information among networks of specialized modules to create higher-level forms of cognition and awareness. We argue that the time is ripe to consider explicit implementations of this theory using deep learning techniques. We propose a roadmap based on unsupervised neural translation between multiple latent spaces (neural networks trained for distinct tasks, on distinct sensory inputs and/or modalities) to create a unique, amodal global latent workspace (GLW). Potential functional advantages of GLW are reviewed, along with neuroscientific implications.

\section{Cognitive neural architectures in brains and machines}
Deep learning denotes a machine learning system using artificial neural networks with multiple ``hidden" layers between the input and output layers.
Although the underlying theory is more than 3 decades old \citep{rosenblatt1958,mcclelland1986}, it is only in the last decade that these systems have started to fully reveal their potential \citep{lecun2015}.
Many of the recent breakthroughs in AI (Artificial Intelligence) have been fueled by deep learning.
Neuroscientists have been quick to point out the similarities (and differences) between the brain and these deep artificial neural networks \citep{kriegeskorte2015,marblestone2016,richards2019,vanrullen2017,yamins2016,lake2017}.
The advent of deep learning has allowed the efficient computer implementation of perceptual and cognitive functions that had been so far inaccessible. Here, we aim to extend this approach to a cognitive framework that has been proposed to underlie perception, executive function and even consciousness: the Global Workspace Theory (GWT).\\

\begin{figure}[ht!]
\centering
\includegraphics[width=0.6\textwidth]{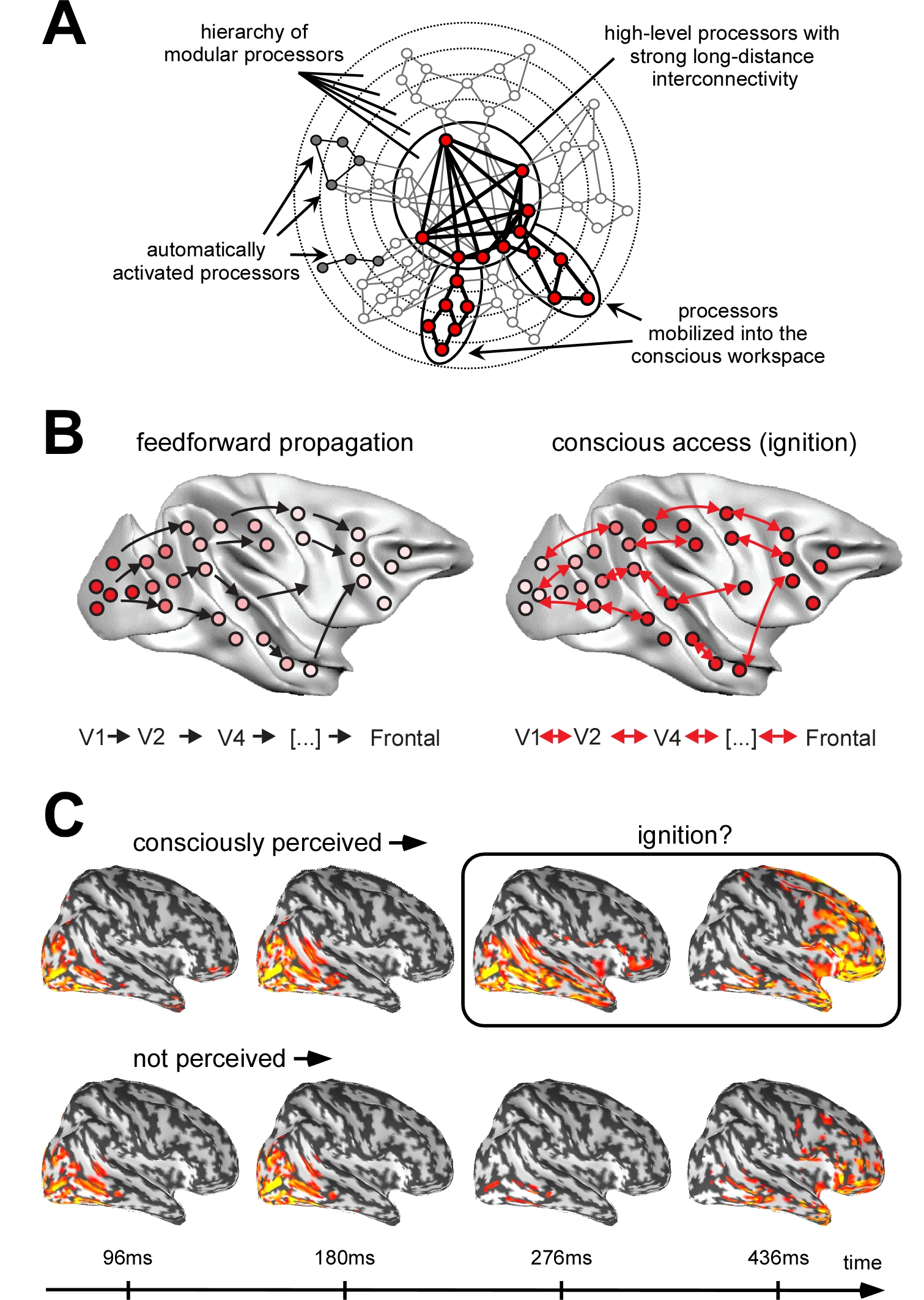}
\caption{\textbf{Global workspace in the brain. A.} Schematic illustration of GWT. Concentric circles depict peripheral (e.g. sensory inputs, motor outputs) vs. more central processes, with the global workspace at the center. Specialized modules process information independently from each other. Their outputs, when selected by bottom-up (saliency-based) or top-down (task-related) attention, can enter the global workspace. There, information processing is characterized by strong long-distance interconnectivity, such that incoming information can be broadcast to other modules. At any given time, a subset of the specialized modules is mobilized into the workspace in a data-dependent and task-dependent manner. The contents of the global workspace reflect our fluctuating consciousness. Redrawn from~\citep{baars1993}. \textbf{B.} Mapping of GWT onto the (monkey) brain. Visual information can propagate through the visual system and activate certain frontal regions controlling behavioral output in a feed-forward way---in this case, information remains unconscious (left). When inputs are sufficiently strong or task-relevant (right), they activate local recurrent connections, resulting in ``ignition" of the global workspace (a highly non-linear, all-or-none process, characterized by global recurrence across a network of long-range connections). Reproduced, with permission, from~\citep{van2018threshold}. \textbf{C.} In certain experimental situations, the same sensory stimulus sometimes reaches consciousness (top row), and sometimes remains unconscious (bottom row). In human magneto-encephalography (MEG) recordings, the main signature of consciously perceived inputs is a late all-or-none activation (or ``ignition'') of frontal regions, accompanied by sustained activity in sensory regions. Adapted, with permission, from~\citep{sergent2005}. } 
\label{fig:GNW}
\end{figure}

The GWT, initially proposed by \citet{baars1993, baars2005global}, is a key element of modern cognitive science (Figure \ref{fig:GNW}A).
The theory proposes that the brain is divided into specialized modules for specific functions, with long-distance connections between them \citep{baars1993, baars2005global}. When warranted by the inputs or by task requirements (through a process of attentional selection), the contents of a specialized module can be broadcast and shared among distinct modules. According to the theory, the shared information at each moment in time---the global workspace---is what constitutes our conscious awareness. In functional terms, the global workspace can serve to resolve problems that could not be solved by a single specialized function, by coordinating multiple specialized modules. \\ 

Dehaene and colleagues \citep{dehaene1998,sergent2004,dehaene2005,van2018threshold,mashour2020} proposed a neuronal version of the theory, Global Neuronal Workspace (GNW), which has become one of the major contemporary neuroscientific theories of consciousness. According to GNW, conscious access occurs when incoming information is made globally available to multiple brain systems through a network of neurons with long-range axons densely distributed in prefrontal, parieto-temporal, and cingulate cortices (Figure \ref{fig:GNW}B). A neural signature of this global broadcast of information is the ignition property: an all-or-none activation of a broad network of brain regions, likely supported by long-range recurrent connections (Figure \ref{fig:GNW}C).\\

Here, we argue that the time is ripe to consider a deep learning implementation of global workspace theory. While Y. Bengio has explicitly linked his recent ``consciousness prior" theory to GWT \citep{bengio2017}, his proposal focused on novel theoretical principles in machine learning (e.g. sparse factor graphs). Our approach is a complementary one, in which we emphasize practical solutions to implementing a global workspace with currently available deep learning components, while always keeping in mind the equivalent mechanisms in the brain. We hope that some of the ideas developed here will assist neuroscientists in interpreting brain data in a new or different light, and in developing novel empirical evaluations of the key operations at play in the global workspace framework. 

\begin{figure}[ht!]
\centering
\includegraphics[trim=0 0 0 0, clip,width=1\textwidth]{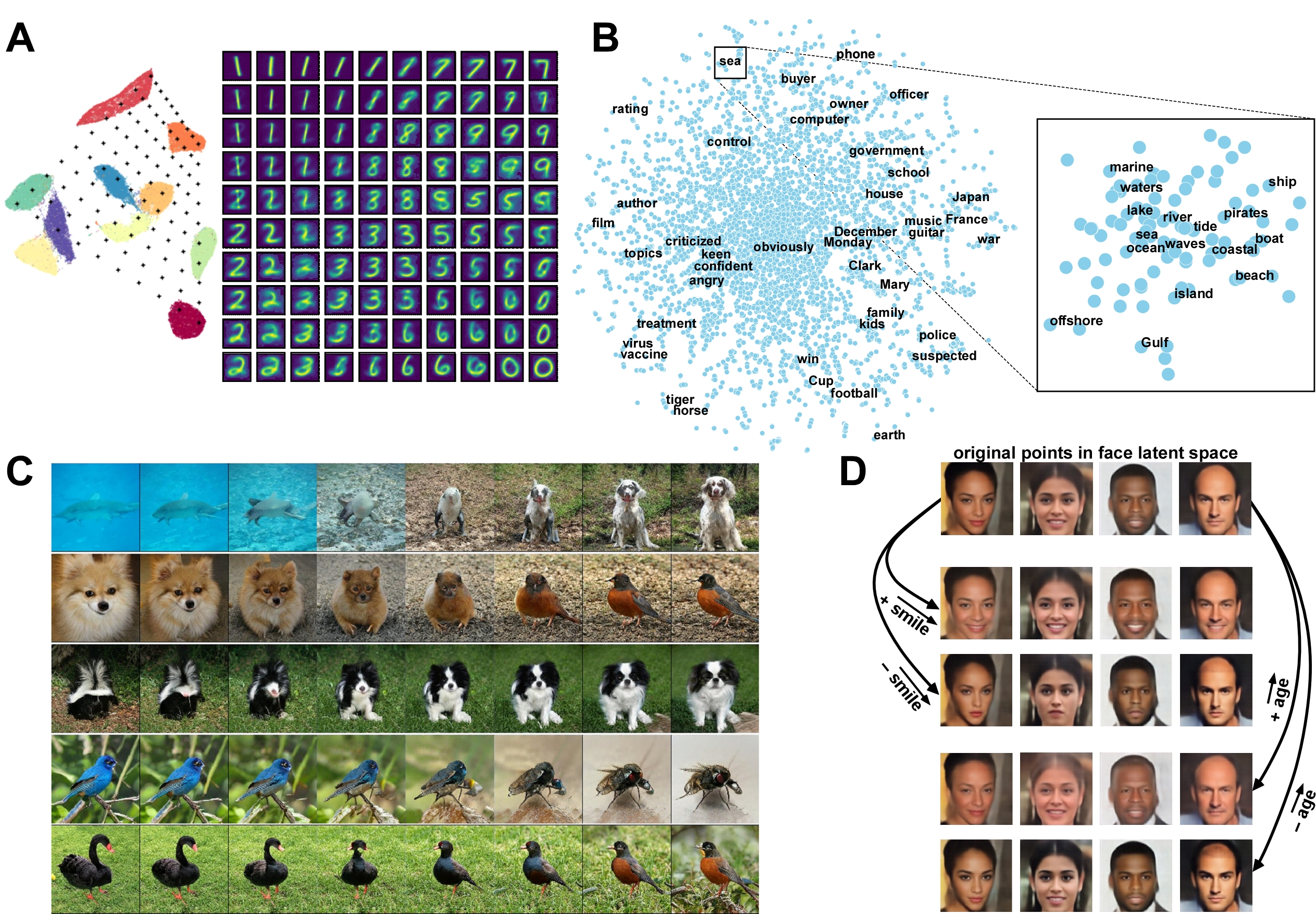}
\caption{\textbf{Examples of deep learning latent spaces:} a low-dimensional space that captures the relevant structure and topology of an input domain or task. In discriminative models, it is often considered to be the last feature layer, and the first layer for generative models. Examples (projected to 2D for visualization) include: \textbf{A.} latent space of the MNIST digit dataset. Each image from the dataset is a point in the space on the left, colored according to digit class. Regularly sampling this space in a 2D matrix produces the image reconstructions on the right (created using the UMAP inverse transform \citep{mcinnes2018}). \textbf{B.} Word embedding space (Word2Vec algorithm \citep{mikolov2013}). Different parts of the latent space focus on distinct semantic domains (e.g. "sea" in the inset). \textbf{C.} Latent space of the ImageNet natural scene dataset derived from the BigGAN generative model \citep{brock2018}. Each row samples different points along a single vector in the 256-D latent space. \textbf{D.} Face latent space from a VAE-GAN model \citep{larsen2016}. In each column, a point is sampled from the latent space, then varying amounts of a pre-computed ``smile" or ``age" vector are added to it. It must be emphasized that latent representations are essentially vectors of neural activation, which can be meaningfully interpolated (as in panels A,C), but also extrapolated and more generally, manipulated with algebraic operations (as in panel D).}
\label{fig:latent}
\end{figure}
\section{Roadmap to a deep learning Global Latent Workspace}
The following is a step-by-step attempt at defining necessary and sufficient components for an implementation of the global workspace in an AI system. Together, these steps define a roadmap towards achieving this goal, and highlight important issues and predictions for neuroscience research. A major point to emphasize is that all of the described components already exist individually, and often reach or surpass human-level performance in their respective functions. The value of our proposal is, therefore, to identify the appropriate components and the manner in which they should interact, so as to optimize functionality while remaining truthful to neuroscience findings. As in any theoretical proposal, some of the details will likely be flawed; in addition, there might be multiple ways to implement a global workspace. Nonetheless, we believe that the strategy outlined below is most likely to be successful.
\begin{itemize}
\item \textbf{Multiple specialized modules.} The first ingredient of GWT is a number ($N \geq 2$) of independent specialized \textbf{modules} (see Glossary), each with their own high-level \textbf{latent space}. In deep learning, a latent space is a representation layer trained to encode the key elements of an input domain. This information corresponds to high-level conceptual representations such as visual object features, word meaning, chunks of action sequences, etc. (Figure \ref{fig:latent}). The modules could be pre-trained neural networks designed for sensory perception (visual or auditory classification, object segmentation...), natural language processing (NLP), long-term memory storage, reinforcement learning (RL) agents, motor control systems, etc. The choice of these specialized modules, of course, is critical since it determines the capabilities of the full global workspace system, and the range of tasks it may perform; however, it does not affect the remaining principles laid out below. \\
\\In theory, connecting together $N$ feed-forward \textbf{discriminative} networks (each trained to classify inputs from their specific domain according to category) could suffice to build a multi-modal workspace (e.g. to preactivate the ``tiger" visual recognition units when one hears the word ``tiger"). In practice, however, there are many reasons why including \textbf{generative} networks would be beneficial---networks that produce motor or language outputs, but also  sensory systems with a generative top-down pathway such as (variational) auto-encoders, GANs or predictive coding networks. This top-down pathway is trivially required if the global workspace is intended to influence the system's behavioral output. It is also necessary (though certainly not sufficient) in order to endow the system with creative or ``imagination" abilities (e.g. generation of mental images), and more generally, to perform mental simulation, planning or ''thinking" by iteratively conjuring up a possible future state or \textbf{counterfactual} state \citep{kanai2019}. Finally, a recurrent top-down pathway may be key to account for the global \textbf{ignition} property observed in the brain, when an input reaches consciousness and the corresponding module is mobilized into the conscious global workspace (Figure \ref{fig:GNW}B,C).

\item \textbf{Global Latent Workspace (GLW).} The GLW, amodal by nature, is an independent and intermediate shared latent space, trained to perform \textbf{unsupervised neural translation} between the $N$ latent spaces from the specialized modules (Figure \ref{fig:workspace}, Key Figure). Although there are numerous examples of supervised multi-modal translation in deep learning \citep{frome2013,desai2020,karpathy2015,silberer2014,qiao2019,kim2018,gorti2018,joseph2019,tsai2019,sun2019,chen2019,harwath2018,li2019,wang2018,pham2019}, here we emphasize \textbf{cycle consistency} as the major unsupervised training objective for neural translation (see Box~1). In brief, the translation system is optimized such that successive translation and back-translation (e.g. a cycle from language A to B, then back to A) always returns the original input. Using this strategy, the GLW can potentially transcribe between any pair of modules, even those for which matched data is unavailable (for example, there is no smell systematically associated with a specific video game state; yet we can intuitively recognize when the player's situation becomes odiferous). Of course, it will be most advantageous if the default unsupervised neural translation strategy can also be complemented by supervised objectives \citep{artetxe2018} whenever joint data is available (e.g. watching an animal while hearing the corresponding sound). The dimensionality of this intermediate space is expected to be on par with or perhaps higher than the dimension of each of the input latent spaces, but much lower than their sum. This bottleneck ensures that only relevant information is encoded at each moment in time, and forces the system to prioritize competing inputs with \textbf{attention}.

\item \textbf{Attention.} In the brain, attention determines what information is consciously perceived, and what is discarded~\citep{posner1994} (although attention and consciousness can be dissociated~\citep{lamme2003visual,koch2007}). Similarly, in the original GWT, attention selects the information that enters the workspace. In deep learning, attention has recently taken the spotlight~\citep{graves2016}, most particularly the transformer architecture used widely in NLP \citep{vaswani2017} and computer vision \citep{wang2017,ramachandran2019,bello2019,zhao2020}. Although the term ``attention'' is the same, there are important differences between the neuroscience and machine learning usage of the notion~\citep{lindsay2020}. In the transformer and related networks, attention is defined as a match between \textit{queries} emitted by one network layer and \textit{keys} produced by another one (possibly the same layer, in the "self-attention" case); the matching score determines what information is passed on to the next stage. Similarly, we can envision a \textit{key-query} matching process to select inputs that reach the GLW and accordingly, to break existing connections or create new ones. If the workspace includes a latent representation of the current task~\citep{zamir2018,achille2019}, this signal can serve to emit a top-down attention \textit{query}, compared against the current ``key'' vectors from all the candidate inputs to the workspace: whenever the latent space of an input module produces a matching \textit{key}, the module is connected and the relevant information is brought into the workspace. In the absence of a clear task, or in the presence of exceptionally strong or surprising inputs, bottom-up attention capture can prevail: in the above terminology, salient information has a ``master key" that supersedes all queries, i.e., that can grant access to the workspace regardless of the current query. The attention mechanism for producing keys and queries in a data-dependent and task-dependent way must be optimized via training with a specific \textbf{objective function} (see Outstanding Questions).  

\item \textbf{Internal copies.} When a specific module is connected to the workspace as a result of attentional selection, its latent space activation vector is copied into the GLW. This \textbf{internal copy} serves the role of a bidirectional connection interface between the corresponding module and the GLW. 

\item \textbf{Broadcast.} The incoming information is then immediately \textbf{broadcast}, that is, translated (via the shared latent space) into the latent space of all other modules. This translation process is automatic: there is no effort involved in consciously apprehending our inner and outer environment. It is how conscious inputs acquire ``meaning", as they suddenly connect to the corresponding linguistic, motor, visual, auditory (etc) representations. This only means that the relevant information in the relevant format is ``available” to these systems (as an internal copy within the workspace), not necessarily that it will be used (i.e., effectively transferred into the corresponding module). One does not always visualize the details of a conjured mental image; one does not always verbalize their thought or inner speech; one does not always act on a motor plan, etc. What determines if this information is used by those systems is whether they are themselves currently connected to the workspace (e.g. by virtue of their task-relevance). The many latent representations that are automatically formed when broadcasting conscious inputs inside the workspace, without being consciously perceived themselves (because their corresponding module is not currently connected to the workspace) may correspond to what Crick and Koch described as the \textbf{penumbra} of consciousness~\citep{crick2003}.

\end{itemize}
\begin{tcolorbox}[floatplacement=h!,float]
\textbf{\\Box 1. Unsupervised neural translation via cycle-consistency.} \\
In Natural Language Processing (\textbf{NLP}), a \textbf{neural translation} system is a machine translation algorithm that uses neural networks. Standard (neural) machine translation is learnt from matched exemplars (words, sentences) in the source and target languages. However, since all languages refer to a common physical reality in the outside world (the so-called \textbf{language grounding} property), their associated semantic representation spaces are likely to share a similar topology: for example, the words ``cat'' and ``dog'' are likely to be found close together, while the word ``machine'' would be more distant (see Figure I). Therefore, it is theoretically possible to learn to align linguistic representations in two (or more) languages based solely on the geometry of their semantic representation spaces, without access to matched corpora (Figure I). This is referred to as \textbf{unsupervised neural translation}. One recently proposed method relies on a \textbf{cycle-consistency} training objective: language alignment is successful when the successive translation from language A to language B, then back from B to A returns the original sentence \citep{he2016, artetxe2018, lample2018}. Similar methods have been applied to neural translation between varied domains, e.g. unpaired image-to-image translation \citep{zhu2017, liu2017,yi2017}, \textbf{text-to-image} translation \citep{chaudhury2017,qiao2019,joseph2019} or \textbf{touch-to-image} translation \citep{li2019}. Domain alignment via cycle-consistency training is also at the heart of a recent surge of studies investigating unsupervised domain adaptation and \textbf{transfer learning} tasks~\citep{hoffman2018,hui2018,murez2018,hosseini2018,tian2019,chen2019cvpr}.\\ 
We suggest that the core challenge for any artificial system based on GWT is in fact a problem of unsupervised neural translation: learning and retrieving appropriate correspondences between elements of distinct domains or modalities, which may not always directly co-occur in the environment. Accordingly, our framework places a strong emphasis on cycle-consistency as an objective function for training the translation mechanism at the heart of the Global Latent Workspace.
\begin{center}
\includegraphics[width=0.9\textwidth]{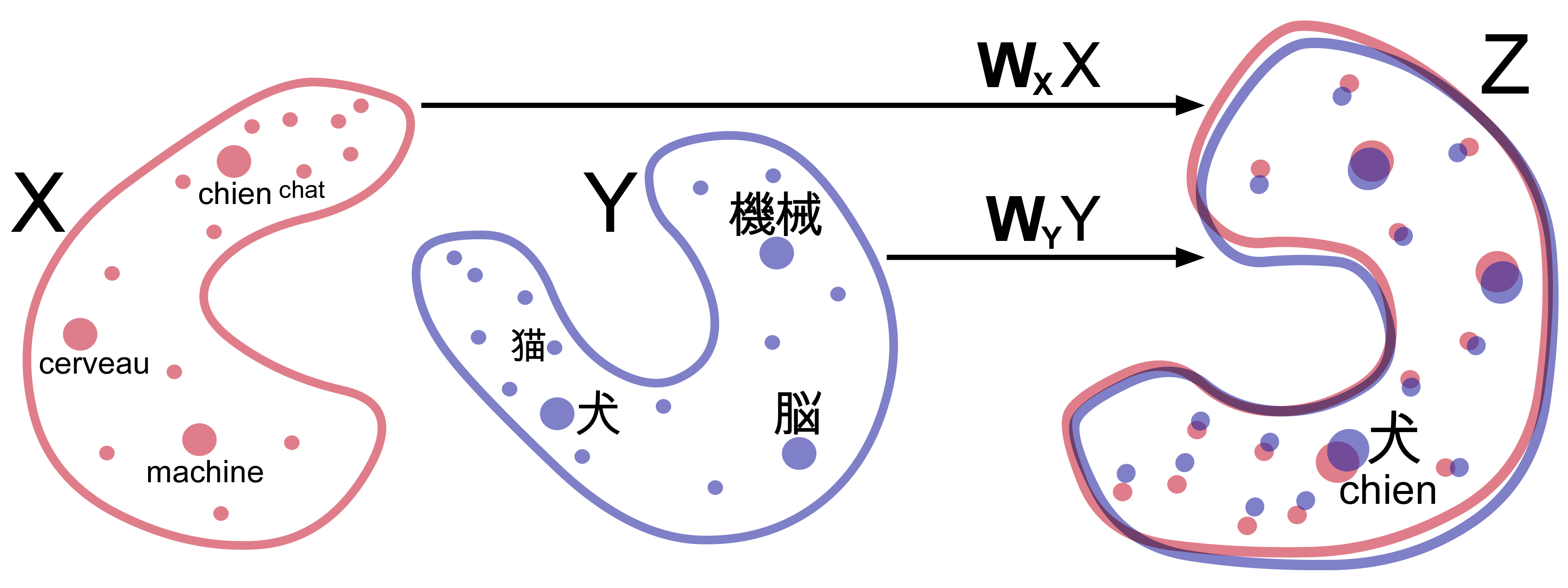}
\end{center}
\small{Figure I. Alignment between linguistic representations. Latent spaces from any two languages $X$ and $Y$ (here, French and Japanese) share a similar topology, and can be aligned to a shared latent space $Z$ through a transform $W$ (adapted from \citep{Conneau2018}).}
\label{box:translation}
\end{tcolorbox}

\section{Global Latent Workspace in action}
\begin{figure}[htbp]
\centering
\includegraphics[trim=0 0 0 0, clip,width=0.75\textwidth]{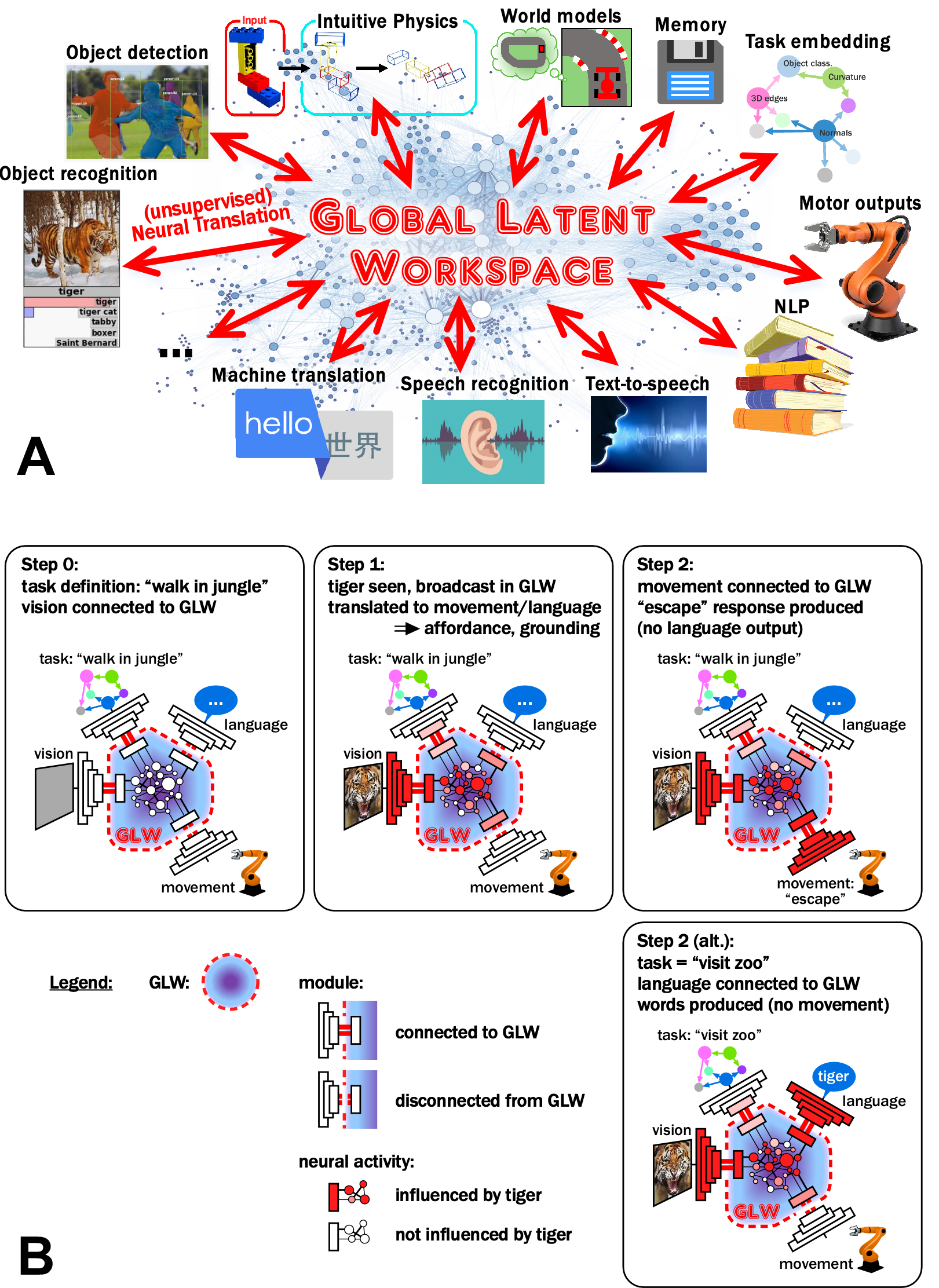}
\caption{\textbf{Schematic of a deep learning ``Global Latent Workspace” (A) and its operation (B).} \textbf{A}. Specialized modules are arranged in the periphery. These can be pretrained networks for any variety of tasks: sensory (object recognition, detection, segmentation, speech recognition...), motor (robotic arm control, speech production...), linguistic (text comprehension, machine translation, text-to-speech...), memory storage, or higher-level cognition- and behavior-related functions (intuitive physics engine, RL policy, task embedding, world model...). Each module is connected to the GLW (schematically represented at the center) via an internal copy of the module's relevant latent space, effectively acting as a connection interface. Through extensive training using a cycle-consistency objective, the workspace learns to translate between the latent space representations of any two modules, in a mostly unsupervised fashion, i.e. without or with very little need for paired data (red arrows). \textbf{B}. When bottom-up or top-down attention (not represented here) selects inputs from one module (Step 0), its latent space activation is copied into the GLW, and immediately translated into representations suitable for each of the other modules (Step 1). However, only a handful of these modules, those currently mobilized into the workspace, will effectively receive and process the corresponding data. For example, upon recognizing a tiger in the visual scene, the corresponding NLP word embedding for ``tiger" and a flight-oriented motor plan would arise in the workspace (Step 1); but the flight would only be initiated (Step 2), or the word ``tiger" pronounced (Step 2alt.), if the corresponding module (motor output, text-to-speech) was effectively recruited in the workspace at this instant.}
\label{fig:workspace}
\end{figure}

To clarify the inner workings of the proposed workspace, let us follow its step-by-step operations during a standard scenario (as illustrated also in Figure \ref{fig:workspace}B). Before any stimulus appears, the prior state of the system, including the current task setting or instructions, can preset some modules to be connected to the GLW, while others remain disconnected (Step 0 in Figure \ref{fig:workspace}B). ``Connected'' means that the latent space of the module is temporarily clamped, in a bidirectional way, to its internal copy in the workspace. If a stimulus appears in a disconnected module, it will not reach the workspace directly; but it may still affect the attentional system (not represented in Figure \ref{fig:workspace}B), which may eventually result in the connection of the relevant module (either because the corresponding key matches the top-down attention query; or because it is a bottom-up ``master key''). If a new stimulus appears in a connected module, the latent activity is immediately transferred to the corresponding ``internal copy'' inside the workspace (Step 1 in Figure \ref{fig:workspace}B). Hence starts the broadcast, that is, an automatic translation to all other domains: via the shared latent space, each internal copy (no matter whether its module is connected or not) receives a translation of the new input in its own ``language''. In turn, the activation from each internal copy will return to the shared latent space and potentially modify it, reverberating ad infinitum inside the GLW. This might be what ``ignition'' means (Figure \ref{fig:GNW}C): long-range and long-lasting recurrent interactions between the latent spaces of the different modules. The shared latent space can use the translations and corresponding ``back-translations'' from all modules to compute its ``cycle-consistency'' error, required to train or fine-tune the unsupervised neural translation system (e.g. via error backpropagation).\\ 
\\ What will modules do with the broadcasted information they receive on their internal copy (Step 2 in Figure \ref{fig:workspace}B)? If the module is disconnected, the broadcast only reaches the internal copy, but not the actual module. Still, this can be helpful for grounding and affordance, as described more fully in the next section. If the module is connected, that is, if its latent space is clamped to the internal copy, the broadcasted information reaching the internal copy will also modify the latent space and potentially the inner layers of the module (for a generative module). For a language network, speech may be produced; for a movement network, an evasive action or a break-dance move may be launched; for a generative visual network, an image reflecting the contents of the workspace may be summoned, etc. This is what it means for a network to be ``recruited'' in the workspace: because of the bidirectional connection, the inner network activity directly affects the GLW, but is also directly affected by activity changes in the workspace.

\section{Functional advantages of a Global Latent Workspace}
A major testable property of the proposed GLW architecture is that the whole should be more than the sum of its parts (i.e., its individual modules). In other words, the added functional properties of GLW, specified below, should result in improved performance across the entire range of modules that are connected to it. Beyond these pre-existing individual tasks (and leaving aside the possible emergence of conscious experience, which we address in the next section), the global workspace also opens up the possibility of combining modules to perform entirely novel tasks.\\

To begin with, the automatic multimodal alignment of representations in GLW is an ideal way to accomplish information \textbf{grounding}. Sensory inputs or motor outputs, instead of meaningless vectors in their respective latent spaces, become associated with corresponding representations in other sensorimotor domains, as well as with relevant linguistic representations: this promotes semantic grounding of sensorimotor data. Conversely with sensorimotor grounding of semantic information, linguistic embedding vectors that merely capture long-range statistical relations between hollow ``language tokens'' are transformed by association with relevant parts of the sensory environment or the agent's motor and behavioral repertoire~\citep{tan2020}. This notion of sensorimotor grounding is thus strongly related to the Gibsonian concept of \textbf{affordance}, and more generally to Gibson's ecological approach in brain science~\citep{gibson1979}. Ultimately, grounded latent representations can confer increased performance to every module connected to the global workspace. We thus predict that GLW should result in performance improvements, particularly in terms of robustness to out-of-distribution samples (including so-called ``adversarial'' attacks~\citep{szegedy2013intriguing}).\\

While grounding and affordance are immediate and automatic consequences of information entering the global workspace, such a system is capable of much more, granted time and effort. Indeed, the ability to transiently mobilize any combination of modules into the workspace in a task-dependent manner is exactly what is required of a general-purpose cognitive architecture. This way, the system can compose more general functions from specialized modules, by deploying one module's abilities onto another module's latent representation. This \textbf{transfer learning} enables agents to adapt to new environments and tasks by generalizing previously learned models, and is considered a core component for implementing intelligence~\citep{legg2007universal,chollet2019measure}. When enough diverse modules are available, their possible combinations are virtually limitless. The price of this flexibility is time and effort: mentally composing functions is a slow, sequential process, requiring iterative calls to top-down attention in order to recruit the relevant modules, one function at a time \citep{sackur2009}. This is what Kahneman, and after him Bengio, have dubbed \textbf{system-2} cognition~\citep{kahneman2011,bengio2017}.\\

One of the major functions that such a flexible mental composition system can produce is \textbf{counterfactual} reasoning, or the ability to answer ``what if?" questions. In this context, a particularly useful module could be a ``world model''. This is an internal model of how the environment reacts to one's actions, which can be queried iteratively as a ``forward model'' to predict future states of the world given an initial state and possible action~\citep{ha2018,hafner2019dream}. This function is at the core of many emblematic attributes of high-level cognition: imagination and creativity, planning, mental simulation, iterative reasoning about possible future states~\citep{kanai2019}.\\ 

Arguably, the cumulative advantages listed here may capture the function of consciousness in humans and animals, as well as a path towards general intelligence in machines.
 
\section{Does GLW entail artificial consciousness ?}
In the original GWT, a necessary and sufficient condition for conscious perception is that the information is broadcast through the global workspace. This raises the question of whether an artificial network equipped with a global latent workspace would necessarily express (a minimal form of) consciousness.
In philosophy of mind and in related neuroscientific theories of consciousness, two aspects of consciousness are usually distinguished~\citep{block1995confusion}: \textbf{phenomenal consciousness} is the immediate subjective experience of sensations, perceptions, thoughts, wants and emotions; \textbf{access consciousness} requires further consolidation, and is used for reasoning and executive control of actions, including language. GWT does not explicitly distinguish between these two forms of consciousness, but other authors have suggested that local recurrence could be sufficient for phenomenal awareness, while global recurrence is a hallmark of access consciousness~\citep{lamme2003visual,lamme2018challenges}. In this view, the global workspace ignition that produces global recurrence of brain activity would more naturally map to access consciousness. Indeed, the functional advantages that we highlight in terms of flexible cognitive control seem in line with the definition of access consciousness, and do not critically depend on the emergence of phenomenal consciousness. Still, there are two aspects of GLW that may be conducive to a form of phenomenal consciousness. First, the grounding and affordance properties could account for the emergence of implicit associations between different sensorimotor properties of an object as well as the associated declarative knowledge (e.g., the word that comes on the tip of the tongue, the different ways we know that we could grasp an object if we decided to, etc). Second, the recruitment of a sensory module into the workspace could explain the vivid and detailed nature of our sensory phenomenal experience: as the connection between the module and its internal copy is bidirectional, the workspace can access sensory information but can also modify it and enrich it with semantically grounded information. Thus, on the one hand, GLW could reasonably be viewed as a way to endow an artificial system with phenomenal consciousness.
On the other hand, our position is that this question is an empirical one, which cannot be addressed without committing to a specific measure of consciousness. The answer, therefore, could heavily depend on the chosen measure: integrated information \citep{tononi2016}, non-trivial information closure \citep{chang2020}, synergistic mutual information \citep{griffith2014}, etc.\\ 
\\ Finally, it is worth noting that the global workspace focuses on the ``information broadcast'' property of awareness. According to \citet{dehaene2017}, there is an additional self-monitoring aspect that is important to capture human and animal consciousness, and that a GLW system as we describe here might be missing. Of course, this self-monitoring itself is likely amenable to a deep learning implementation, but we defer this question to future work.

\section{Implications for Neuroscience}
A global workspace using unsupervised neural translation to broadcast information between internal copies of every modality-specific latent space---if it exists in the brain---should have a number of telltale signatures that could be explored by neuroscientists. An internal copy, for example, would correspond to a population of neurons having a clear preference for a specific sensorimotor domain or modality, but whose response is heavily influenced by high-level, semantic or multimodal information (the grounding and affordance properties). While there are many candidate high-level or multimodal regions in the brain, the concept of internal copy further implies that the activation of this neural population (i) could happen without stimulation of its preferred modality, yet (ii) would be systematically coupled to a global ignition of the entire workspace. \\
\\ Training the translation mechanism by optimizing cycle-consistency may be relatively straightforward to implement with biological neurons, by making the networks mutually predictive of each other. In this sense, cycle-consistency could be envisioned as a form of ``predictive coding'', a well-studied framework in neuroscience~\citep{rao1999predictive,bastos2012canonical}. Broadcast implies a recurrent loop between translations, back-translations and error estimations, resembling the prediction error minimization objective of predictive coding. As this sort of error minimization loop is also known to be a source of brain oscillations~\citep{bastos2012canonical,alamia2019alpha}, we further suggest that internal copy neurons in the brain could be characterized by oscillatory responses at a specific frequency. \\
\\ A dedicated attention system is required to control the workspace inputs and outputs. In our framework (inspired by the deep learning transformer architecture~\citep{vaswani2017}), the workspace constantly emits context-dependent attention queries, each module emits attention keys, and the match between keys and queries determines the module's connection status. In the brain, this would correspond to endogenous attention systems, particularly the dorsal part of the frontoparietal network responsible for top-down attention control~\citep{corbetta2002control}. As we explained, it would be advantageous if the modules with especially salient inputs had the ability to emit ``master keys'' to force their recruitment into the workspace, regardless of the current query. This is a form of bottom-up attention capture, reminiscent of the ``circuit-breaking'' property of exogenous attention in the brain~\citep{corbetta2002control,itti2001computational}. Finally, while the workspace requires a dedicated and unified attention system, this does not preclude the existence of other independent attention systems within each module. Similarly in the brain, there are global forms of attention to select entire modalities while inhibiting others, but also more ``local'' forms of attention operating within each modality, e.g. to highlight one object among others~\citep{macaluso2002directing}. For optimal performance, these multiple attention systems should be allowed to interact~\citep{driver1998attention}, for instance by sharing queries. The resulting widespread network of within- and between-modality attention systems could correspond to the so-called frontoparietal attention network~\citep{corbetta2002control,szczepanski2013functional}. \\ 
\\ Common neuronal or cognitive phenomena may be revisited in the light of our proposed framework. For instance, the suppression of consciousness during general anesthesia has been linked to a specific impairment of long-range connections~\citep{mashour2013cognitive,mashour2020}, which are crucial for the normal operation of the global workspace---specifically for ignition, broadcast and translation. A model like the one we propose could serve to perform pre-clinical studies, e.g. to evaluate how various aspects of consciousness depend on certain anesthetic drug targets. Just like anesthesia may reflect impaired translation mechanisms, synesthesia could be related to hyperactive translation. Humans show the ability to discover patterns through analogical reasoning, as well as a natural tendency to connect seemingly unrelated stimuli in a consistent manner--a tendency that culminates in the arbitrary and mandatory cross-modal associations of synesthesia~\citep{hubbard2005neurocognitive}. Yet neuronal mechanisms for such phenomena have been elusive. The unsupervised neural translation we discussed here offers a possible algorithmic method to establish such alignment of high-level representations across modalities, and could thus help understand the origins of synesthesia.

\section{Concluding remarks}
Having a roadmap towards GLW does not imply that this goal is easy to reach---actual implementation will involve much trial-and-error, and  as yet unknown computational resources. Of course, it is not the first time that a computer implementation of GWT is suggested~\citep{franklin2006,shanahan2006cognitive,bengio2017,bao2020,safron2020,kotseruba2020}. What sets our stance apart is the conjunction of two factors. First, we capitalize on modern deep learning-compatible components, most of them validated in state-of-the-art neural network architectures. Second, we contemplate the underlying neuronal bases and the neuroscientific implications of the proposed scheme. Correspondingly, we hope that this work may serve two purposes. Firstly, from a cognitive neuroscience standpoint, considering how to effectively implement the global workspace theory forces us to be very concrete about each component of the theory, and thereby gives us an opportunity to refine the corresponding notions. In turn, these refined notions could help formulate new hypotheses that may be empirically tested using neuroscientific methods. Secondly, in the context of artificial intelligence, the main implication of our effort is to show that inspiration from neuro-cognitive architectures may have important functional benefits. GLW could serve to improve specific machine learning tasks or benchmarks by augmenting existing architectures, thanks to the added robustness conferred by the grounding of representations inside the workspace. But GLW could also be a way to develop entirely novel architectures capable of planning, reasoning and thinking through the flexible reconfiguration of multiple existing modules. This may bring us one step closer to general-purpose (system-2) artificial cognition.\\
\begin{tcolorbox}
\textbf{Outstanding questions}
\begin{itemize}[itemsep=2pt, leftmargin=*]
\item A global workspace serves to flexibly connect neural representations arising in multiple separate modules. Is there a minimal number of modules feeding into the workspace? When does bimodal, trimodal, multimodal integration become a “global workspace”?
    
\item Can we identify neurons, e.g. in frontal regions, that incarnate copies of the various latent spaces? This may explain the numerous reports of sensory and multimodal neuronal responses in frontal cortex.

\item Is cycle-consistency implemented in the brain? If yes, does it correspond to a form of predictive coding? 

\item Could \textit{synesthesia} be the consequence of an exaggerated or overactive translation between domains, crossing the threshold of perception instead of acting as a background process?    

\item How does attention learn to select the relevant information to enter the GLW? What is the corresponding objective function? Many candidates exist and could be tested: self-prediction, free energy, survival, reward of a RL agent, metalearning (learning progress), etc.

\item How can newly learned tasks or modules be connected to an existing GLW? Requirements include: a new ``internal copy" with a new (learned) attention mechanism to produce keys for the latent space, new (learned) translations to the rest of the workspace.
\end{itemize}
\label{box:outstanding}
\end{tcolorbox}
\begin{tcolorbox}[breakable]
\textbf{Glossary}
\\Our terminology is borrowed from different fields, with the same term sometimes taking distinct meanings across the fields. To alleviate any confusion, we begin each definition by indicating whether the term is employed in a way traditionally associated with Cognitive Neuroscience \textit{(Neuro)} or AI \textit{(AI)}.
\begin{itemize}[itemsep=2pt, leftmargin=*]
\item \textbf{affordance:} \textit{(Neuro)} objects and events are interpreted according to the options they offer an observer in terms of available uses (including mental usage) and possible actions: their \textit{affordances}
\item \textbf{attention:} \textit{(AI)} bottom-up or top-down selection of information to enter the workspace, by means of matching \textit{query} and \textit{key} vectors
\item \textbf{broadcast:} \textit{(AI)} automatic translation of incoming information from one selected module into a format suitable for the latent space of all other modules
\item \textbf{counterfactual:} \textit{(Neuro/AI)} resulting from simulation of possible situations, without a direct connection to reality or facts
\item \textbf{cycle-consistency:} \textit{(AI)} objective function for translation between two domains A and B, whereby successive translations from A to B and from B back to A should retrieve the original input
\item \textbf{discriminative/generative network:} \textit{(AI)} a neural network in which information flows from the external environment towards the latent space is called \textit{discriminative}, and \textit{generative} for the opposite direction; some networks can be both (with bidirectional information flow)
\item \textbf{grounding:} \textit{(Neuro)} how representations from one domain acquire ``meaning", by associating them with other related (and possibly unrelated) domains
\item \textbf{internal copy:} \textit{(AI)} the GLW contains an internal copy of each module's latent space, used for automatic translation and broadcast; recruiting a module into the workspace amounts to effectively connecting this internal copy to the corresponding latent space
\item \textbf{latent space:} \textit{(AI)} low-dimensional space that captures the structure and topology of an input and/or output domain (for discriminative or generative networks, respectively) 
\item \textbf{module:} \textit{(AI)} a specialized system, operating independently of the GLW, but capable of connecting to it when needed (to achieve this, the module's latent space gets clamped to its internal copy in the workspace)
\item \textbf{neural translation:} \textit{(AI)} machine translation algorithm that uses neural networks
\item \textbf{objective function:} \textit{(AI)} the measure that a network aims to optimize via training
\item \textbf{penumbra:} \textit{(Neuro)} according to Crick and Koch, the ensemble of neural activity produced by the current conscious state, yet not strictly part of it
\item \textbf{phenomenal/access consciousness:} \textit{(Neuro)} the immediate subjective experience of sensations, emotions, thoughts (etc.) is called \textit{phenomenal} consciousness; \textit{access} consciousness denotes information used for reasoning and executive control of actions, including language 
\item \textbf{supervised/unsupervised learning:} \textit{(AI)} training a network with/without a desired output corresponding to each input 
\item \textbf{system-2:} \textit{(Neuro/AI)} cognitive architecture capable of deliberate planning and reasoning, typically slow and effortful compared to immediate perceptual awareness, well-practiced tasks or reflexive behaviors
\item \textbf{transfer learning:} \textit{(AI)} application of a model trained on one problem to a distinct but related problem. Domain adaptation tasks are a subset of transfer learning
\end{itemize}
\label{box:glossary}
\end{tcolorbox}
\begin{tcolorbox}
\textbf{Highlights}

\begin{itemize}[itemsep=2pt, leftmargin=*]
\item In recent years, deep learning has steadily improved the state-of-the-art in artificial intelligence, but mainly for single, well-defined tasks or challenges
\item Novel advanced neural network architectures, possibly inspired by Neuroscience, are needed to create more general-purpose AI systems with flexible and robust capabilities
\item The 30-year old Global Workspace Theory proposed such an architecture; we now consider its implementation in a deep learning framework
\item The Global Workspace accounts for conscious information processing in the human brain, but its associated functional advantages could generalize to artificial systems
\item In turn, considering an artificial global workspace can help constrain neuroscientific investigations of brain function and consciousness

\end{itemize}
\label{box:highlights}
\end{tcolorbox}
\section*{Acknowledgments}
RV is supported by an ANITI (Artificial and Natural Intelligence Toulouse Institute) Research Chair (grant ANR-19-PI3A-0004), and two ANR grants AI-REPS (ANR-18-CE37-0007-01) and OSCI-DEEP (ANR-19-NEUC-0004). RK is supported by Japan Science and Technology Agency (JST) CREST project. We wish to thank Leila Reddy, Thomas Serre, Andrea Alamia, Milad Mozafari and Benjamin Devillers for helpful comments on the manuscript.

\printbibliography

\end{document}